\documentclass[11pt,letterpaper]{article}
\usepackage{emnlp2016-tex/emnlp2016}
\usepackage{times}
\usepackage[hyphens]{url}
\usepackage{latexsym}
\usepackage{array}
\usepackage{amsmath}
\usepackage{amssymb}
\usepackage{xspace}
\usepackage{algorithm}
\usepackage[noend]{algpseudocode}
\usepackage{color}
\usepackage{scalefnt}
\usepackage{microtype}
\usepackage{booktabs}
\usepackage{graphicx}
\usepackage{linguex}

\newcommand{\tc}{Object-Only\xspace}
\newcommand{\bw}{Object+Attribute\xspace}

\emnlpfinalcopy

\def\emnlppaperid{***}

\title{``Show me the cup'': Reference with
  Continuous Representations}

\author{Gemma Boleda$^*$ \ Sebastian Pad\'o$^\dagger$ \  Marco Baroni$^*$\\
$^*$Center for Mind/Brain Sciences\\
University of Trento\\
\texttt{firstname.lastname@unitn.it}\\
$^\dagger$Institut f\"ur Maschinelle Sprachverarbeitung\\
Universit\"at Stuttgart\\
\texttt{sebastian.pado@ims.uni-stuttgart.de}}

\date{}

\begin{document}

\maketitle

\begin{abstract}
    One of the most basic functions of language is to \textit{refer} to
  objects in a shared scene. Modeling reference with continuous
  representations is challenging because it requires
  \textit{individuation}, i.e., tracking and distinguishing an
  arbitrary number of referents. We introduce a neural network model
  that, given a definite description and a set of objects represented
  by natural images, points to the intended object if the expression
  has a unique referent, or indicates a failure, if it does not. The
  model, %
 directly trained on reference acts, is competitive with a
  pipeline manually engineered to perform the same task, both when
  referents are purely visual, and when they are characterized by a
  combination of visual and linguistic properties.

\end{abstract}

\section{Introduction}
\label{sec:introduction}

\begin{figure}[h]
\center
\includegraphics[width=7cm]{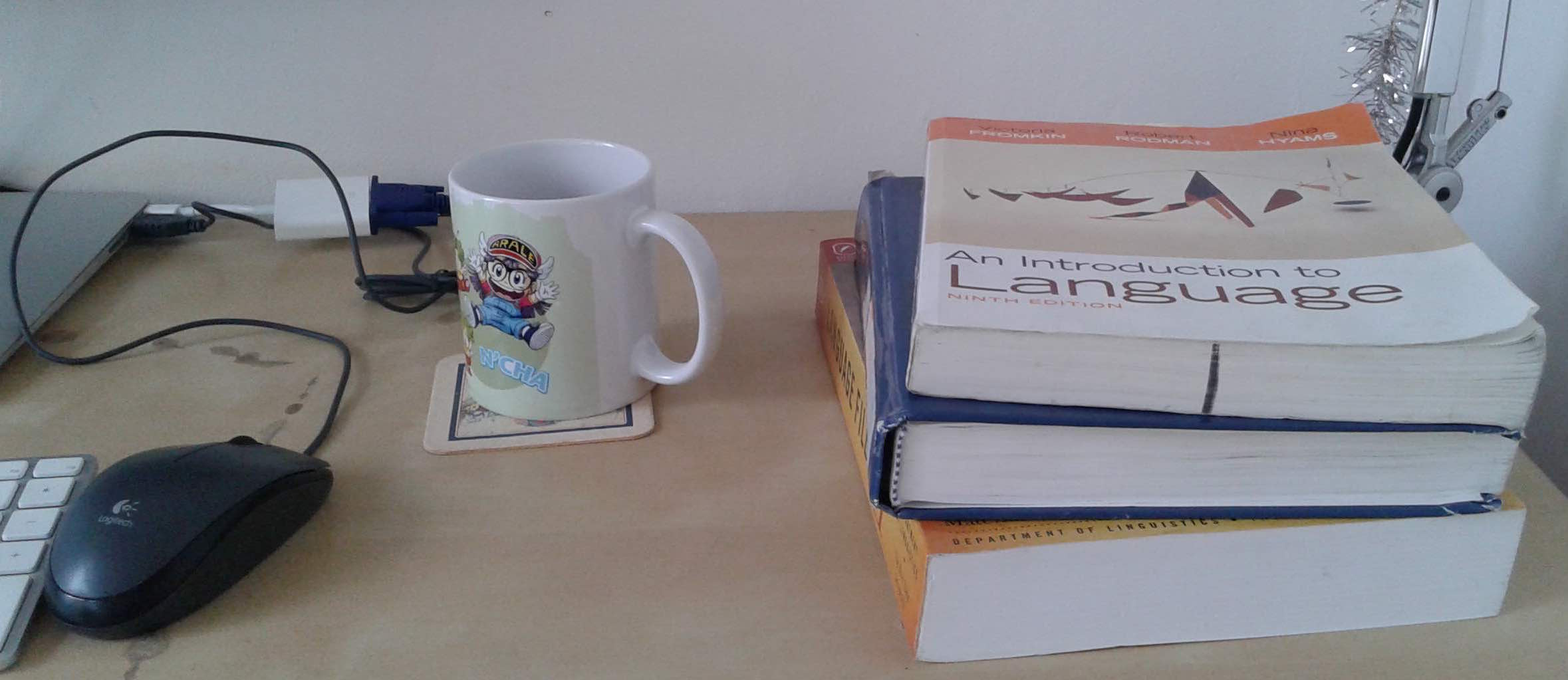}
\caption{Example scene.}
\label{fig:natural-scene}
\end{figure}

Humans use language to talk about the world, and one of its most basic
functions is to \textit{refer} to objects
\cite{russell05:_denot}. This makes reference one of the fundamental
devices to \textit{ground} linguistic symbols in extralinguistic
reality \cite{Harnad1990}.
\footnote{We ignore the thorny philosophical
  issues of reference, such as its relationship to reality. For an
  overview and references (no pun intended), see
  \newcite{sep-reference}.}  
For successful reference, the speaker
must choose an expression allowing the hearer to pick the right
referent. For instance, assume that Adam and Barbara are in the
context of Figure~\ref{fig:natural-scene}, and consider the
 dialogues in~\ref{ex:ref-examples}.

\vspace{-0.2cm}
\ex.\label{ex:ref-examples} \emph{Adam}: Can you please give me\dots \vspace{-0.1cm}
\a.\label{ex:a} \dots the mug?\\
\emph{Barbara}: Sure.
\b.\label{ex:b} \dots the pencil?\\
\emph{Barbara (searching)}: Ahem, I can't see any pencil here\dots
\b.\label{ex:c} \dots the book?\\
\emph{Barbara}: Sorry, which one?

In dialogue~\ref{ex:a}, reference is successful. It fails in
\ref{ex:b} and \ref{ex:c}, but for different reasons: in \ref{ex:b},
the word ``pencil'' does not apply to any object in the scene; in
\ref{ex:c}, the use of singular ``the'' implies that Adam refers to a
unique object, while the scene contains three matching objects. These
examples show how reference involves both
\emph{characterization} mechanisms that capture object properties,
mainly through the use of content words (e.g.\ ``mug'' vs.\
``pencil''), and \emph{individuation} mechanisms, prominently encoded
in function words and morphology (e.g., ``the'' vs.\ ``some'',
singular vs.\ plural), which allow us to track and distinguish
referents. 


Existing computational approaches to meaning account for one of these
aspects at the expense of the other: Data-driven approaches, including
distributional semantic and neural network models, typically model the
conceptual level \cite{Turney2010}, accounting well for
characterization, but not for individuation. The converse holds for
logics-based approaches \cite{Bos:2004:WSR:1220355.1220535}.




In this paper, we propose a neural network model aimed at both aspects
of reference, and that can be trained directly on reference acts. Just
like in the typical reference scenario, the model works across
modalities, looking for the referent of a verbal expression in the
visual world, or in a setting in which entities are characterized by
joint visual and linguistic information. The model,
\textit{Point-or-Protest (PoP)}, behaves like Barbara: It identifies
(\textit{points} to) the image that corresponds to a given linguistic
expression, or \textit{protests} in case of reference failure. While
the model is generic and could be extended to other reference types,
our starting point in this paper is reference to (concrete) entities
using single-entity denoting noun phrases (as in
\ref{ex:ref-examples}). This case clearly illustrates the joint
workings of characterization (reference requires recognizing the right
sort of entity in the scene) and individuation (reference succeeds
only if there is \emph{exactly one} entity of the right kind: in
\ref{ex:ref-examples}, Barbara cannot simply recognize the presence of
some ``pencil mass'', but she must check that there is only one pencil
to unambiguously refer to). %
We show, in two experiments, that PoP is competitive with a
state-of-the-art pipeline requiring specific heuristics.\footnote{We will
  make our code and data available.}



\section{Models}
\label{sec:models}

\begin{figure*}[t]
\center
\includegraphics[width=\textwidth]{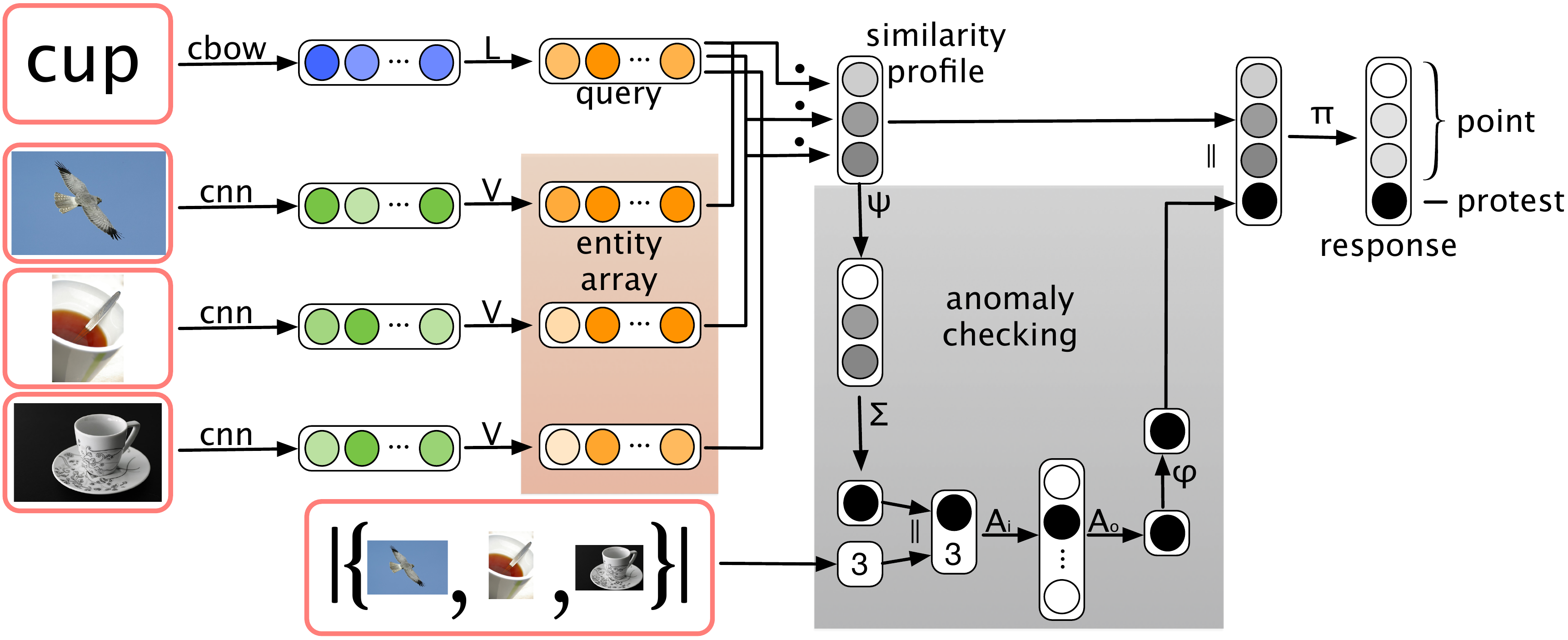}
\caption{\textbf{The point-or-protest (PoP) model}.  Network inputs
  are marked with salmon-pink frames. Uppercase Latin letters
  represent linear transformations and lowercase Greek letters
  nonlinearities. $\|$ stands for vector concatenation, the period represents
  dot products, $\Sigma$ stands for summation across the values of a
  vector. Vectors containing ellipses may have different
  dimensionality than the one depicted; among such vectors, those with
  the same color belong to the same space (have the same number of
  dimensions). The intensity of a cell's fill is informally meant to
  express the size of the value it contains.}
\label{fig:model-diagram}
\end{figure*}

\paragraph{Point-or-Protest} Point-or-Protest (PoP) is a feed-forward
neural network learning from examples how to react to successful and
failed reference acts.\footnote{For neural network design and training
  see, e.g., \newcite{Nielsen:2015}.} Given a variable-length sequence
of objects depicted in images (possibly coupled with other information
characterizing them, e.g., verbal attributes) and a natural language
query, PoP must either \emph{point} to the object denoted by the
query, returning its index in the sequence, or \emph{protest} if the
query phrase is not an appropriate referring expression. The PoP
architecture builds an ``entity array'' whose entries are vectors
storing information about the objects in the scene, and uses
similarity-based reasoning about the vectors in the array and the
query to decide its response. We currently focus on singular definite
article semantics, as in~\ref{ex:ref-examples}, with failure if there
is no possible referent (missing-referent anomaly) or if there is more
than one (multiple-referent anomaly). We discussed above the
linguistic appeal of this case. From a machine-learning perspective,
one-entity individuation requires a non-linear separation of the
anomalous reference acts (0 or more than 1) from the felicitous ones.

We use the diagram in Figure \ref{fig:model-diagram} to introduce
PoP. In this example, the input set contains a harrier and two cups,
with the corresponding linguistic query being \emph{cup}.\footnote{We
  do not enter the determiner in the query, since it does not vary
  across data points: our setup is equivalent to always having ``the''
  in the input. The network learns the intended semantics through
  training.} PoP should thus raise the anomaly flag.

The linguistic query is first mapped to a dense space by using pre-compiled
\texttt{cbow} embeddings, whereas images are mapped to vector
representations by passing them through a pre-trained convolutional
neural network (\texttt{cnn}), and extracting the activation patterns
in one of the top layers of the network (see Section
\ref{sec:experiments} for further details). If
the input consists of objects with linguistic attributes, we simply
concatenate the corresponding \texttt{cnn} and \texttt{cbow} vectors
to get their input representation, and analogously we concatenate
\texttt{cbow} vectors to represent multi-word linguistic phrases. %
Conceptually, using \texttt{cbow} embeddings means that the
listener we model already possesses large amounts of unembodied
knowledge about word meaning, as gathered from linguistic co-occurrence
patterns independently of reference. This assumption is unrealistic,
and we abandon it with the TRPoP model described below.

PoP maps the input object representations in the sequence to an array
of entity vectors by applying a linear transformation. The
corresponding mapping matrix $\mathbf{V}$ is shared across objects, as
the position of objects in the input sequence is arbitrary, and PoP
should not learn associations between objects and specific sequence
slots (e.g., from the Figure~\ref{fig:model-diagram} example, it
should not learn to associate cups with positions 2 and 3 in
general). Each vector in the entity array corresponds to one input
object. In parallel, PoP maps the linguistic expression to a ``query''
vector through a separate linear transformation $\mathbf{L}$. The
query vector lives in the same space as the entity vectors to enable
pairwise similarity computations. We can thus interpret the matrices
$\mathbf{V}$ and $\mathbf{L}$ as mapping input vectors into a shared
multimodal space, in which it is possible to probe visual (or mixed)
entities with linguistic queries. Next, the network takes the dot
product of the query with each entry in the entity array. The
resulting vector (containing as many dimensions as dot products, and
thus objects) encodes the similarity profile of the query with the
entity vectors: the larger the value in dimension $n$, the more likely
it is that the $n$-th object in the input sequence is a good referent
for the query.

PoP also needs to assess whether the reference act was felicitous. The
cumulative ``similarity mass'' across entity vectors should provide
the network with good evidence to reason about anomaly. For the
specific aim of modeling singular reference, the network should
discover that, when cumulative similarity is too low or too high, the
reference is not appropriate for the current sequence: in the first
case, because no object matches the query; in the second, because
there is more than one object that matches the query. More precisely,
along the ``anomaly pathway'' shown in grey in Figure
\ref{fig:model-diagram}, we first pass the similarity vector through a
nonlinearity $\psi$ to sharpen the contrasts, particularly zeroing out
low similarities. For example, a \texttt{relu} transformation might
set all low similarities to 0, making it easier to detect anomalies:
in Figure~\ref{fig:model-diagram}, the whitening of the harrier
similarity cell is meant to suggest this process. We then sum across
all values in the resulting vector, obtaining a cumulative similarity
score. We concatenate it with the \textit{cardinality} of the input
sequence and feed them, via a linear transformation $\mathbf{A}_i$, to
a vector of ``anomaly sensor'' cells. Cardinality enables the model to
take the number of inputs into account when assessing the cumulative
similarity score: the same score that looks suspiciously high for two
objects is bound to be low for ten objects. More specifically, through
cardinality the model can compare the average similarity to arbitrary
thresholds, and subsets of anomaly sensor cells can learn different
thresholds to pick up anomalies (the presence of multiple anomaly
sensor cells allows the model to pick up ``non-linear'' patterns, such
as the one for single-entity reference we are addressing here). Their
output is linearly combined via matrix $\mathbf{A}_o$ into a single
value. The latter is passed through nonlinearity $\phi$, that is
bounding the anomaly score to approximate a discrete yes/no response.

We finally concatenate the similarity profile with the cell containing
the anomaly score, and pass the resulting vector through a
\texttt{softmax} nonlinearity ($\pi$). The model output for an input
sequence of $n$ objects will thus contain a probability distribution
over $n+1$ indices. We take the index with the maximum value for this
distribution as PoP's response: if it is one of the first $n$ indices,
then PoP ``pointed'' at the corresponding object, whereas if PoP
assigned maximum probability to the $n+1$th cell, that means that it
``protested''. 
 In the
figure diagram, PoP has correctly raised the anomaly flag. 
We train PoP by backpropagating the error of the log-likelihood cost
function when comparing its output (either the index of the
correct object, or the anomaly flag) with the ground truth for the
training reference acts.


\begin{figure*}[t]
  \centering
  \includegraphics[width=\textwidth]{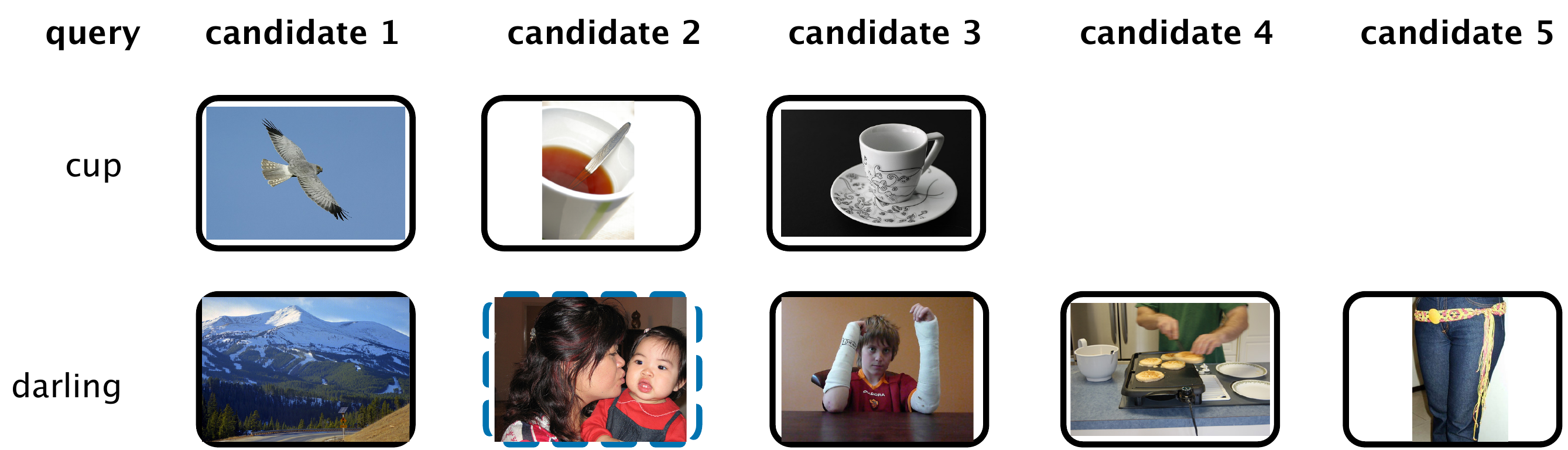}
  \caption{Example sequences from the \tc dataset: Multi-referent
    anomaly (top) and successful reference (bottom, correct image marked with
    blue dashed frame).}
  \label{fig:tc}
\end{figure*}

\paragraph{Pipeline} As a strong competitor, we implemented a method
that performs our task by manual pipelining of a set of separately
trained/tuned components. The Pipeline first induces a set of
multimodal embeddings by optimizing similarities between matched pairs
of queries and objects, compared to random confounders. It uses a
max-margin cost function forcing query representations to be (much)
more similar to the objects they denote than to irrelevant ones. This
has been shown to produce excellent multimodal embeddings
\cite{Frome:etal:2013,Lazaridou:etal:2015d,Weston:etal:2011}. Once
these embeddings have been separately trained, the model computes
similarities between the query and each of the objects in each
referential act in our test sets, picking the object with largest
similarity as candidate object to point at. Then, two separately-tuned
heuristics are used to catch anomalous acts: Missing reference is
predicted if no query-object similarity is above an (optimized)
threshold. Multiple reference is guessed if the difference between the
two largest similarities is below another optimized threshold.


\paragraph{Convolutional Neural Network} Since PoP uses input image
embeddings based on a pre-trained convolutional neural network (CNN),
we also test a model matching the categorical labels produced by the
same CNN for the input images against the query. For the example of
Figure~\ref{fig:model-diagram}, it would pass each of the images
through the full CNN, obtaining 3 labels. We take a lax approach to
label matching, in which the model scores a hit even when, e.g., the
gold label is a substring of the model-predicted one. Anomaly
detection is straightforward (although again implemented ad hoc): CNN
deems a reference act anomalous if no produced label matches the
query, or if more than one does. Thus, the CNN would be successful if
it predicted a synonym of \textit{cup} for both image 2 and 3.

\paragraph{Tabula Rasa PoP} Through the \texttt{cbow} vectors, PoP can
rely on pre-acquired text-induced word similarity knowledge. %
The assumption that word meanings are first learned separately, purely
from language statistics, and then fine-tuned in the referential
setup, is unrealistic. Ideally, we would want a model that learns word
representations in parallel from reference acts \emph{and} language
statistics. For the time being, we consider instead the other extreme,
where word representations are entirely induced from the reference
acts during training. The ``Tabula Rasa'' PoP model (TRPoP) is
identical to the one in Figure\ \ref{fig:model-diagram}, except that
input query representations (and attributes in the \bw setup explained
below) are one-hot vectors. This model will thus induce distributed
representations from scratch when estimating the weights of matrices
$\mathbf{L}$ and $\mathbf{V}$. Such representations will then depend
entirely on the role of words as queries or attributes in the
referential acts we model.

\section{Data}
\label{sec:data}

We test our model in two experiments, for each of which we have
automatically created a large-scale dataset. Both datasets contain
40,000 sequences for training, 5,000 for validation and 10,000 for
testing, each with 15\% missing-referent and 15\% multi-referent
anomalies. The sequences are of varying length, from 2 to 5 candidate
referents. The supplementary materials contain the algorithms used to
generate the datasets as well as detailed statistics.



\paragraph{\tc Experiment.} Our first experiment represents a base
case of reference, namely matching noun phrases consisting of single
nouns with visually represented entities. Figure~\ref{fig:tc} shows
two examples. 
The objects and images are sampled uniformly
at random from a set of 2,000 objects and 50
ImageNet\footnote{\url{http://imagenet.stanford.edu/}} images per
object, itself sampled from a larger dataset used in Lazaridou et
al.~\shortcite{Lazaridou:etal:2015b}. As the examples 
show, we use natural objects and images, which makes the task
very challenging (even humans might wonder which image in the
second row depicts a \emph{darling}).  We generate data with an algorithm sampling sets of sequences with uniform
distributions over sequence lengths (2 to 5) and indices of the
queried object within a sequence.

\begin{figure}[t]
  \includegraphics[width=\columnwidth]{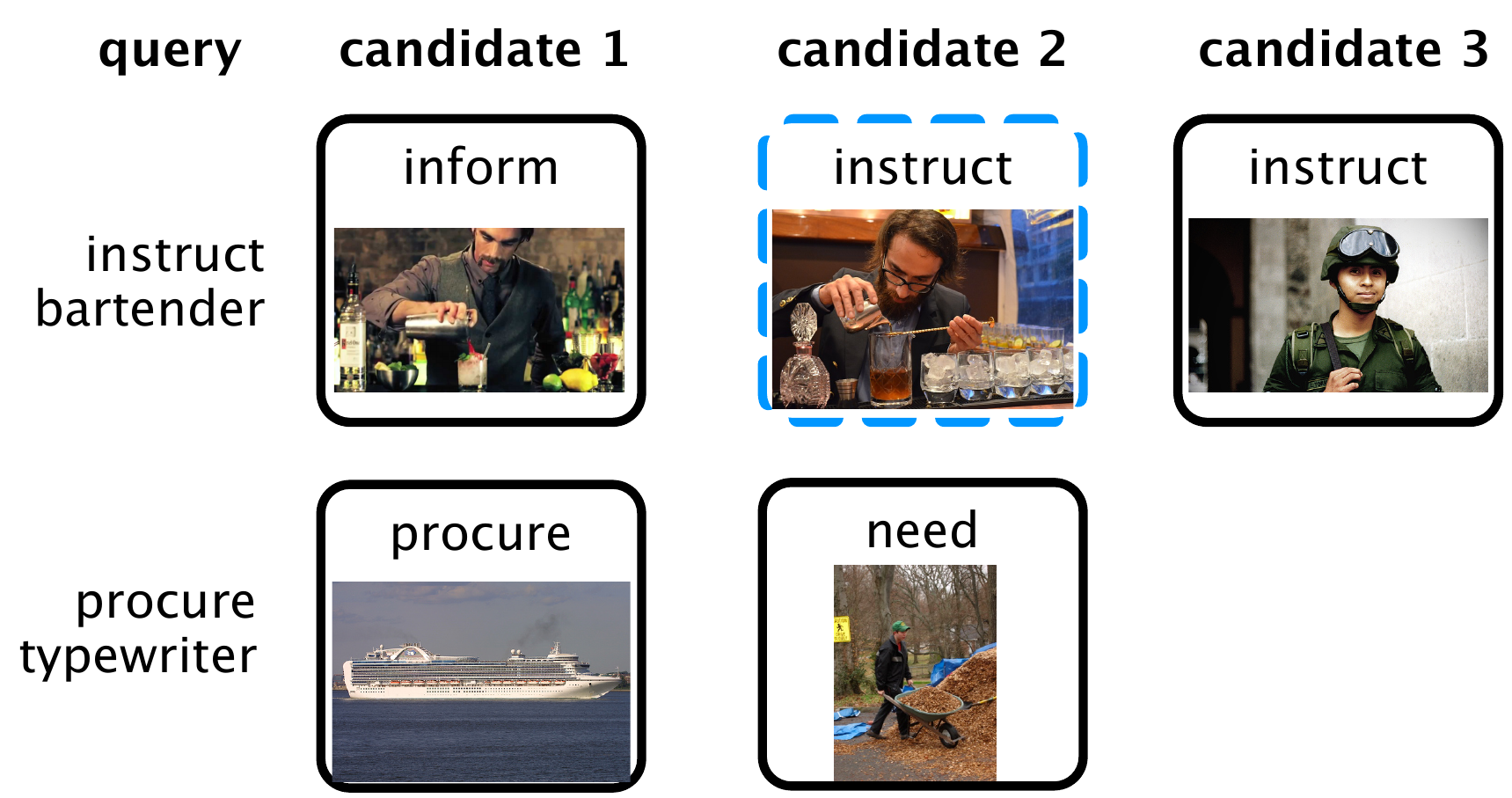}
  \caption{Short \bw example sequences: successful reference (top) and
    missing-referent anomaly (bottom).}
  \label{fig:bw}
\end{figure}

\paragraph{\bw Experiment.} Our second experiment, illustrated
in Figure~\ref{fig:bw}, goes one step
further in testing the model's individuation capabilities.  In the
scene from Section~\ref{sec:introduction}, imagine that Adam points to
the book on top and says ``I recommend this book''. This
linguistically conveyed information will be associated to Barbara's
representation of the entity, together with its visual
features. Crucially, it can be used to identify the first book if
later on Adam asks her ``Can you bring the book I recommended?''. We
test this situation in a simplified form. Each referent is associated
with both an image and a linguistically-expressed \textit{attribute},
more specifically a verb (the only word class from which we could
sample a sufficient number of attributes with the characteristics
outlined below). The query and the sequence items are all pairs like
\textit{spend:bill}, where we interpret the attribute analogously to
an object relative clause, that is, a \textit{bill} that is being
\textit{spent} (we ignore tense for simplicity). 
%

We restrict the attributes under consideration for each object to the
500 highest-associated syntactic neighbors of the object according to
the DM resource\ \cite{Baroni2010}, such that the attributes be
compatible with the objects (to exclude nonsensical combinations
such as \textit{repair:dog}). Of these, we retain only verbs taking
the target item as direct object, in line with the ``relative clause''
interpretation sketched above. Moreover, we focus on (relatively)
abstract verbs, for two reasons. First, 
a concrete verb is more likely
than an abstract one to have strong visual correlates that do not
match what is actually depicted in an image (cf.~\emph{groom:dog}
vs.~\emph{like:dog}). Second, successful reference routinely mixes
concrete and abstract cues (e.g., a noun referring to a concrete
object combined with a modifier recording an event associated to it:
\emph{the book I lent you}), and we are interested in
simulating this scenario. We thus filter verbs through the
concreteness norms of Brysbaert et
al.~\shortcite{brysbaert14:_concr_englis}, retaining only those with a
concreteness score of at most 2.5 (on a 1--5 scale).

The object-attribute structure of the stimuli in this experiment also
enables us to introduce challenging confounders into the sequences --
namely, pairs that share either the attribute or the object with the
query. 
For each sequence, we start by picking an attribute-object
query. Given the query, we generate two more compatible attributes for
the query object, and alternative objects compatible to these
attributes as well as the initial attribute. Starting from all
attribute-object combinations,~we randomly drop as many as necessary
to obtain the final sequences of 2 to 5 items.  A consequence of this
design is that the objects within sequences tend to be somewhat
related since they share compatible attributes, and vice versa. The
first sequence in Fig.~\ref{fig:bw} illustrates the effect: For the
query object \textit{bartender}, we generate the confounder object
\textit{soldier}, connected through the attributes \textit{instruct}
and \textit{inform}. The full sequence also includes the confounder
object \textit{emperor}, not shown in the figure.


\section{Experiments}
\label{sec:experiments}

\begin{table*}[tb]
  \centering
  	\begin{small}
  \begin{tabular}{lcccccccc}
    \toprule
    & \multicolumn{4}{c}{Exp 1: Object Only} & \multicolumn{4}{c}{Exp
      2: Object+Attribute}\\
    \cmidrule(lr){2-5} \cmidrule(lr){6-9}
             &Total&Pointing&MissRef&\multicolumn{1}{c}{MultRef}&Total&Pointing&MissRef&MultRef\\
    \midrule
    PoP           &66   &71      &57     &51                         &69   &77      &57     &46\\
    TRPoP         &65   &70      &58     &44                         &62   &70      &38     &48\\
    Pipeline      &67   &75      &51     &45                         &65   &74      &37     &55\\
    CNN           &35   &9       &100    &94                         &-   &-      &-     &- \\
    \midrule
    Random        &17      &17         &17        &17                &17   &17      &17     &17\\
    Majority      &30      &0          &100       &100               &30   &0       &100    &100\\
    Probability   &22      &18         &30        &30                &22   &18      &30     &30\\
    \midrule
    AttrRandom    & -  & -     & -    & -                            &47   &64      &16     &0\\
    ImgShuffle    & -  & -     & -    & -                            &50   &58      &31     &32\\
    \bottomrule
  \end{tabular}
  \caption{\textbf{Results.} Figure of merit is percentage accuracy. See text for details.}
  \label{tab:all-results}
   \end{small}
\end{table*}

\paragraph{Method} PoP and Pipeline's input word representations are 400-dimensional
\texttt{cbow} embeddings from \newcite{Baroni:etal:2014}, trained on
about 2.8 billion tokens of unannotated text. These models, as well as
TRPoP, use 4096-dimensional vectors as input visual representations,
which are produced by passing images through the pre-trained VGG
19-layer CNN of \newcite{Simonyan:Zisserman:2015} (trained on the
ILSVRC-2012 data), and extracting the corresponding activations on the
topmost fully connected layer.\footnote{We use the MatConvNet toolkit,
  \url{http://www.vlfeat.org/matconvnet/}} The same pre-trained
network was used to generate the labels of our CNN competitor
model. The parameters of PoP/TRPoP and of the Pipeline max-margin
embeddings are estimated by online stochastic gradient descent on the
training portions of the two datasets. For Pipeline, we extract all
possible pairs of positive and negative query-object tuples from each
reference act in the relevant training data. Details on model
hyperparameter tuning are in the supplementary material. We consider
three baselines for both experiments. \textbf{Random} assigns all
labels randomly. \textbf{Majority} assigns the most frequent output
label, namely anomaly, accounting for 30\% of the sequences (the
non-anomalous labels are distributed among predicted
indices). \textbf{Probability} randomly assigns labels based on their
relative frequency in the training data.





\paragraph{Experiment 1: Object-Only} Results are reported on the
left-hand side of Table \ref{tab:all-results}. Besides overall
accuracy (\emph{Total}), we show accuracy itemized by successful
reference acts (\emph{Pointing}), missing-referent (\emph{MissRef})
and multiple-referent anomalies (\emph{MultRef}).

(TR)PoP and Pipeline are clearly above the baselines (Majority and
Probability reach deceptively high anomaly-detection scores by
over-raising the anomaly flag, at the cost of pointing
performance). PoP's absolute performance is close to that of the
manually-crafted Pipeline. By jointly learning to point and handling
anomalies in reference acts, PoP loses some performance in pointing,
but in exchange it does better on anomaly detection. As could be
expected, MissRef is more difficult than MultRef for all three
models. Interestingly, TRPoP, which does not rely on pre-trained word
embeddings, performs comparably to PoP (but it requires more than
twice as many epochs to converge, see supplementary materials). This
suggests that useful representations of word meaning can be learned
solely from examples of successful and failed reference acts.

CNN performance is barely above baseline, and, like Majority and
Probability, it trivially reaches high performance on anomaly cases
because it raises the anomaly flag whenever it fails to produce the
name of the target object (and it rarely produces the right
label). For instance, in the first example in Figure~\ref{fig:tc}, CNN
can get a hit as long as it doesn't produce ``cup''. As for its
extremely low pointing performance, note that CNN, unlike PoP, cannot
make reasonable pointing guesses for objects it did not see during
training. The large performance asymmetry between these two models
sharing the same visual processing network shows that PoP generalizes
well beyond the knowledge it inherited from this pre-trained
network. Importantly, PoP reasons about similarity in multimodal
space, rather than assigning hard labels. For example, the CNN, when
presented with an image associated to the out-of-training query
\emph{academician}, tags it as \emph{academic gown} -- not
unreasonably but incorrectly. PoP points to the correct slot because
its multimodal \emph{academician} query vector is most similar to the
correct entity vector than to the other candidates, with no need to
perform explicit label matching. Intriguingly, even when considering
the subset of test data that CNN is trained on, we still observe an
asymmetry: CNN reaches 58\% accuracy, while PoP's performance is at
67\%. This suggests that reference-based training has fine-tuned
better representations also for the objects the CNN was explicitly
trained for.

\paragraph{Experiment 2: Object+Attribute} Results are shown on the
right side of Table \ref{tab:all-results}. CNN is not tested here, as
it does not handle attributes. PoP's results are slightly higher than
in the previous experiment, while those of TRPoP and Pipeline are
slightly lower, such that now PoP is clearly above them. The three
models are exploiting both visual and verbal information, as shown by
their comparison to two additional baselines, shown at the bottom of
the table. AttrRandom randomly picks one of the objects that shares
the attribute with the query, if any, and raises the anomaly flag
otherwise. This baseline has, by construction, 0\% accuracy on MultRef
anomalies, and it performs at random in MissRef detection. However,
even in the pointing case, its performance is still well below that of
the models.  ImgShuffle is a variant of PoP trained after shuffling
image vectors, so that each image ID is (consistently) associated with
the CNN representation of another image (mostly depicting objects that
do not match the image label). The only reliable signal that this
baseline can then exploit is attribute information. Again, its
performance is clearly below that of the models.

As for anomaly handling, while PoP still finds MissRef easier than
MultRef, this time TRPoP and Pipeline actually perform worse on
MissRef. Comparing the MissRef cases in which the models failed to
raise the anomaly flag, we observe that Pipeline and TRPoP wrongly
pointed to an entity sharing the query attribute much more often than
PoP (943 and 883 vs.~629). They are thus over-relying on matching
attributes, assigning too high a similarity to pairs that are simply
sharing the verbal attribute. This also explains their higher
performance on MultRef: Attribute sharing makes both repeated
referents very similar to the query, triggering the relevant
heuristic. 
Thus, PoP seems better at integrating verbal and visual cues than
them. Compared to TRPoP, PoP has an important prior in the semantics
encoded in its pre-trained word embeddings, which helps it discover
systematic relations between words and objects while keeping the
attribute information apart from that of the head noun. Compared to
Pipeline, by jointly learning to point and to spot anomalies, it might
be able to attain a better balance between visual and verbal
information.



To conclude, our model PoP and its variant TRPoP can learn to refer
directly from examples. While PoP is not clearly superior to Pipeline,
it has a fundamental advantage: It learns to refer in one integrated
architecture. Pipeline (as well as CNN) learns to \emph{characterize}
objects (e.g., to recognize cups as referents for ``cup''), but uses
an ad hoc strategy, needing a manually coded heuristic, to simulate
\emph{individuating} capabilities (distinguishing cases where there
are several or no cups). As soon as the referential setup gets more
complex, as in the \bw experiment in which visual and verbal
information need to be combined, the heuristics break down.

\section{Related work}
\label{sec:related}

\paragraph{Modeling.} The PoP model ``reasons'' about the similarity
between a query and a set of candidates in vector space, akin to soft
attention mechanisms in recent neural network architectures
\cite{Bahdanau:etal:2015,Xu:etal:2015}. While attention is standardly
used to retrieve auxiliary information when producing an output, we
directly expose the similarity vector as (part of the) output, in
order to obtain a model that learns to point robustly across input
sequence orders and lengths. The idea of exposing an attention
mechanism functioning as a pointer over the input has recently been
employed by \newcite{Vinyals:etal:2015b} in the context of
sequence-to-sequence RNNs. PoP's entity array emulates traditional
memory locations within a fully differentiable architecture. This is
akin to the memory vectors of the recently proposed Memory Networks
framework
\cite{Sukhbaatar:etal:2015,Weston:etal:2015a}. 
However, the Memory Networks array has fixed size, whereas our entity
array adapts to input object cardinality.

\paragraph{Multimodal reference resolution.} Our task is a special
case of reference resolution. Various studies in this area have
proposed multimodal approaches jointly handling vision and language
\cite[a.o.]{Gorniak:Roy:2004,Larsson:2015,Matuszek:etal:2014,Steels:Belpaeme:2005}. These
papers focus on aspects of the resolution process we are not currently
addressing, such as full compositionality or gesture, but they work
with very limited perceptual input, such as simple shapes and
colours. Probably the most relevant study in this area is the one by
Kennington and Schlangen \shortcite{Kennington:Schlangen:2015}. They
consider visual scenes with more objects than our sequences, but more
limited in nature (tables with 36 puzzle pieces). They handle spatial
relations and flexible compositionality. However, they must train a
separate classifier for each word in their set, which means that their
method can't process unseen words, and would probably perform badly
for words that are not observed frequently enough during
training. Moreover, they do not present an integrated architecture for
the whole resolution process, as we do, but separate components that
are manually combined. Crucially, they assume referring expressions
are always felicitous. We are not aware of prior work that, like our
Object+Attribute setup, considers referents disambiguated by a mixture
of perceptual and verbally-expressed abstract properties.

\paragraph{Referring expression generation and other related work.}
The task of referring expression generation \cite{J12-1006} has
recently received new impulses from the study of multimodal
language/vision scenarios. The task is converse to ours: given a
scene, generate the optimal linguistic expression to pick out a given
referent. The focus is generally on considerably more complex (but
artificial or heavily controlled) scenes than our sequences, and
correspondingly on linguistically more complex referring
expressions. Some recent efforts collect and analyze large corpora of
referring expressions for multimodal tasks
\cite{DBLP:conf/emnlp/KazemzadehOMB14,tily2009refer}.
A method to generate unambiguous referring expressions for objects in
natural images has been recently proposed by
\newcite{Mao:etal:2016}. Our task is more distantly related to visual
question answering
\cite{Geman:etal:2015,Malinowski:Fritz:2014,Ren:etal:2015}, in the
sense that we model one specific type of question that could easily be
asked about an image. %
Even more generally, our approach fits into the multimodal
distributional semantics paradigm. See \newcite{Baroni:2016} for a
discussion of how the problem of reference is addressed in that line
of work. There is of course a large body of work on modeling reference
with symbolic/logical methods \cite{Abbott:2010}, that provides the
framework for our problem, but is not directly relevant to our
empirical aims. %
Our task can finally also be seen as a special case of the much
broader problem of content-based image retrieval
\cite{Datta:etal:2008}.

\section{Conclusion and outlook}

PoP is a neural network model that, given a linguistic expression and
a set of objects represented by natural images, either resolves
reference by pointing to the object denoted by the expression, or
flags the reference act as anomalous if the linguistic expression is
not adequate. The model consists of an integrated and generic
architecture that can be directly trained on examples of successful
and failed reference acts. The model is competitive with a pipeline
manually engineered to perform the same task. PoP successfully
accounts for characterization because it is able to relate entity
properties (visual, and linguistically conveyed) to linguistic
expressions, via its distributed representations. It has some
individuating capabilities because it builds entity representations
and reasons about them.

Although our experiments concentrated on singular definite
descriptions (``\emph{show me THE cup''}, the PoP architecture is
general enough that should be able, given appropriate training data,
to learn to respond to other reference acts, e.g., corresponding to
\emph{all X} or \emph{many X}~\cite{sorodoc+16}.  We intend to
pursue this direction in future work. We also intend to progressively
remove various artificial characteristics of our simulations, such as
the fact that our sequences do not form natural scenes, and that our
linguistic expressions are very limited. Most importantly, we are
currently assuming one-to-one mapping between images and entity
vectors. In real-life reference, however, possible referents might
re-appear or be mentioned at different times and in different places,
and we might, over time, acquire further knowledge characterizing
them. PoP must thus eventually be able to learn when to initialize a
new entity vector (thus maintaining distinct, \emph{individuated}
representations of the entities in ongoing discourse), and when to
update an existing one with new information (furthering the
\emph{characterization} of the entity). Recent work on fully
differentiable architectures controlling discrete data structures that
can grow and shrink, such as stacks \cite{Joulin:Mikolov:2015}, should
make such extensions feasible.



\section*{Acknowledgments}

We are grateful to Elia Bruni for the CNN baseline idea, and to
Angeliki Lazaridou for providing us with the visual vectors used in
the paper. This project has received funding from the European Union's
Horizon 2020 research and innovation programme under the Marie
Sklodowska-Curie grant agreement No 655577 (LOVe); ERC 2011 Starting
Independent Research Grant n.~283554 (COMPOSES); DFG (SFB 732, Project
D10); and Spanish MINECO (grant FFI2013-41301-P).

\bibliography{refs,marco}
\bibliographystyle{emnlp2016-tex/emnlp2016}

\end{document}


\maketitle

This supplementary file formally defines the algorithms for the
creation of the two datasets we use (Sections 1 and 2), provides
statistics on the final datasets (Section 3) and on hyperparameter
tuning (Section 4).

\section{Data Creation for the \tc Dataset (Experiment~1)}

The process to generate a object sequence is shown in Algorithm
\ref{alg:create-thecat}. We start with an empty sequence and sample
the length of the sequence uniformly at random from the permitted
sequence lengths (l.\ 2). We fill the sequence with objects and
images sampled uniformly at random (l.\ 4/5).  We assume, without loss
of generality, that the the object that we will query for, $q$, is
the first one (l.~6).  Then we sample whether the current sequence
should be an anomaly (l.\ 7).  If it should be a missing-anomaly
(i.e., no matches for the query), we overwrite the target object and
image with a new random draw from the pool (l.\ 9/10). If we decide to
turn it into a multiple-anomaly (i.e., with multiple matches for the
query), we randomly select another position in the sequence and
overwrite it with the query object and a new image
(l.~12/13). Finally, we shuffle the sequence so that the query is
assigned a random position (l.\ 14).

\begin{algorithm}[ht]
  \caption{Creation of \tc dataset}
  \small
  \label{alg:create-thecat}
  \begin{algorithmic}[1]
    \Require Sequence length interval $[i \geq 2,j]$; Set of objects
    $O=\{o_1,\dots,o_n\}$ and sets of associated images $I(o)$ for
    each object $o$; probability of missing-anomalies $P_0$;
    probability of multiple-anomalies $P_m$.

    \Ensure $\langle$ object query $q$, object-image sequence $S$$\rangle$

    \State $S \gets []$
    \State $l \sim \mathit{U}(i,j)$ 
    \For {$k=1$ to $l$}
  \State $o \sim O, i \sim I(o)$
  \State $S[k] = \langle o,i\rangle$ 
    \EndFor
    \State $q \gets S[1]$
    \State $r_0 \sim \operatorname{Bern}({p_0})$, $r_m \sim \operatorname{Bern}({p_0+p_m})$
    \If{$r_0$} 
    \State $o' \sim O, i' \sim I(o)$ so that $o' \not= q$
    \State $S[0] \gets \langle o,i\rangle$
    \ElsIf{$r_m$}
    \State $i \sim \mathit{U}(1,l)$
    \State $S[i] \gets \langle o,i'\rangle$ where $S[1]=\langle o,i
    \rangle$, $i' \sim I(o)$
    \EndIf
    \State \text{shuffle}($S$)
  \end{algorithmic}
\end{algorithm}

\section{Data Creation for the \bw Dataset (Experiment~2)}

\begin{figure}[t]
  \centering
  \includegraphics[width=0.8\columnwidth]{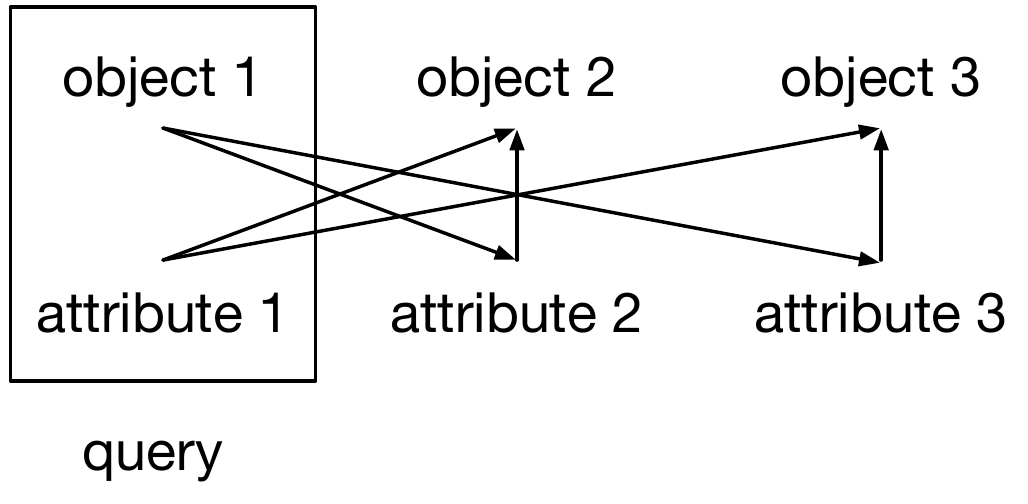}
  \caption{Sampling intuition for \bw}
  \label{fig:bw-image}
\end{figure}

\begin{table*}[t]
  \centering
  \small
  \begin{tabular}{l rrrr rrrr rrrr}
    \toprule
     & \multicolumn{4}{c}{Train set avg.~frequency} & \multicolumn{4}{c}{Test set
       avg.~frequency} & \multicolumn{4}{c}{unseen in test set (\%)}  \\
     \cmidrule(lr){2-5} \cmidrule(lr){6-9} \cmidrule(lr){10-13}
    & O & O+I  & O+A & O+A+I & 
     O & O+I  & O+A & O+A+I & 
     O & O+I  & O+A & O+A+I \\
    \midrule
    \tc & 90.0 & 2.0 & -- & -- 
        & 22.5 & 1.2 & -- & --  
        & 0.0 & 23.1 & -- & -- \\
    \bw & 90.9 & 2.2 & 8.2 & 1.1 
        & 23.1 & 1.3 & 2.7 & 1.0 
        &  0.0 & 20.2 & 0.9 & 82.9  \\
    \bottomrule
  \end{tabular}
  \caption{Statistics on \tc and \bw datasets. O: object, A: attribute, I: image.}
  \label{tab:data-stats}
\end{table*}

Figure~\ref{fig:bw-image} shows the intuition for sampling the \bw
dataset. Arrows indicate compatibility constraints in sampling.  We
start from the query pair (object 1 -- attribute 1). Then we sample
two more attributes that are both compatible with object 1. Finally,
we sample two more objects that are compatible both with the original
attribute 1 and one of the two attributes.

Algorithm~\ref{alg:create-biologistwoman} defines the sampling
procedure formally. We sample the first triple randomly (l.\ 2). Then
we sample two two compatible attributes for this object (l.\ 3), and
one more object for each attribute (l.\ 4). This yields a set of six
confounders (l.\ 5--10). After sampling the length of the final
sequence $l$ (l.\ 11), we build the sequence from the first triple and
$l-1$ confounders (l.\ 12--13), with the first triple as query (l.\
14). The treatment of the anomalies is exactly as before.

\begin{algorithm}[tb]
  \caption{Creation of \bw dataset}
  \small
  \label{alg:create-biologistwoman}
  \begin{algorithmic}[1]
    \Require Sequence length interval $[i \geq 2,j]$; Set of objects
    $O=\{o_1,\dots,o_n\}$, sets of associated
    images $I(o)$ and associated abstract attributes $A(o)$ for each object
    $o$; probability of missing-anomalies $P_0$; probability of

    multiple-anomalies $P_m$.
    \Ensure $\langle$object-attribute query $q$, object-image-attribute sequence $S$$\rangle$

    \State $S \gets []$, $S_c \gets []$
    \State $o_1 \sim O, a_1 \sim A(o_1), i_1 \sim I(o_1)$
    \State $a_2,a_3 \sim A(o_1)$ so that $a_1 \not= a_2 \not=a_3$
    \State $o_2 \sim A^{-1}(m_2)$, $o_3 \sim A^{-1}(m_3)$
    \State $S_c[1] \gets \langle a_2, o_1, i \sim I(o_1) \rangle$
    \State $S_c[2] \gets \langle a_1, o_2, i \sim I(o_2) \rangle$
    \State $S_c[3] \gets \langle a_2, o_2, i \sim I(o_2) \rangle$
    \State $S_c[4] \gets \langle a_3, o_1, i \sim I(o_1) \rangle$
    \State $S_c[5] \gets \langle a_1, o_3, i \sim I(o_3) \rangle$
    \State $S_c[6] \gets \langle a_3, o_3, i \sim I(o_3) \rangle$
    \State $l \sim \mathit{U}(i,j)$ 
    \State $S[1] \gets \langle o_1, a_2, i_1 \rangle$
    \State $S[2..l] \gets$ sample candidates from $S_{c}$ w.o. replacement
    \State $q \gets S[1]$
    \State $r_0 \sim \operatorname{Bern}({p_0})$, $r_m \sim \operatorname{Bern}({p_0+p_m})$
    \If{$r_0$} 
    \State $o' \sim O, a' \sim A(o), i' \sim I(o)$ so that $\langle o',a'\rangle \not= q$
    \State $S[0] \gets \langle a', o', i \rangle$
    \ElsIf{$r_m$}
    \State $i \sim \mathit{U}(1,l)$
    \State $S[i] \gets \langle a,o,i'\rangle$ where $S[1]=\langle a,o,i
    \rangle$, $i' \sim I(o)$
    \EndIf
    \State \text{shuffle}($S$)
  \end{algorithmic}
\end{algorithm}

\section{Statistics on the Datasets}


Table~\ref{tab:data-stats} shows statistics on the dataset. The first
line covers the \tc dataset. Objects occur on average 90 times in the
train portion of \tc, specific images only twice; the numbers for the
test set are commensurately lower. While all objects in the test set
are seen during training, 23\% of the images are not. Due to the
creation by random sampling, a minimal number of sequences is repeated
(5 sequences occur twice in the training set, 1 four times) and shared
between training and validation set (1 sequence). All other sequences
occur just once.

The second line covers the \bw dataset. The average frequencies for
objects and object images mirror those in \tc quite closely. The new
columns on object-attribute (O+A) and object-attribute-image (O+A+I)
combinations show that object-attribute combinations occur relatively
infrequently (each object is paired with many attributes) but that the
combination is considerably restricted (almost no combinations are new
in the test set). The full entity representations
(object-attribute-image triples), however, are very infrequent
(average frequency just above 1), and more than 80\% of these are
unseen in the test set. A single sequence occurs twice in the test
set, all others once; one sequence is shared between train and test.

\section{Hyperparameter Tuning}

We tuned the following hyperparameters on the Object-Only validation
set and re-used them for Object+Attribute without further tuning
(except for the Pipeline heuristics' thresholds). Chosen values are
given in parentheses.
\begin{itemize}
\item  \textbf{PoP}: multimodal embedding size
(300),  anomaly sensor size (100), nonlinearities $\psi$ (\texttt{relu}) and $\phi$
  (\texttt{sigmoid}), learning rate (0.09), epoch count (14).
\item \textbf{TRPoP}: same settings, except epoch count (36). 
\item \textbf{Pipeline}: multimodal embedding size (300),
  margin size (0.5), learning rate (0.09), maximum similarity
  threshold (0.1 for \tc, 0.4 for \bw), top-two similarity difference
  threshold (0.05 and 0.07). 
\end{itemize}
Momentum was set to 0.09, learning rate decay to 1E-4 for all models,
based on informal preliminary experimentation.